\pdfoutput=1

\documentclass[11pt]{article}

\usepackage{acl}

\usepackage{times}
\usepackage{latexsym}

\usepackage[T1]{fontenc}

\usepackage[utf8]{inputenc}

\usepackage{microtype}

\usepackage{inconsolata}

\usepackage{graphicx}
\usepackage{framed}
\usepackage{subcaption}
\usepackage{CJKutf8}
\usepackage{float} 
\usepackage{enumitem} 
\usepackage{multirow} 
\usepackage{booktabs} 
\usepackage{array}
\title{Beyond English-Centric LLMs: What Language Do Multilingual Language Models Think in?}

\author{
Chengzhi Zhong\textsuperscript{1}\hspace{1em}
Fei Cheng\textsuperscript{1} \hspace{1em} 
Qianying Liu\textsuperscript{2} \hspace{1em}
Junfeng Jiang\textsuperscript{3} \hspace{1em} \\
\textbf{Zhen Wan\textsuperscript{1} } \hspace{1em}
\textbf{Chenhui Chu\textsuperscript{1} } \hspace{1em} 
\textbf{Yugo Murawaki\textsuperscript{1} } \hspace{1em} 
\textbf{Sadao Kurohashi\textsuperscript{1,2} }\\
\textbf{\textsuperscript{1}} Kyoto University, Japan \hspace{1em}
\\
\textbf{\textsuperscript{2}} National Institute of Informatics, Japan \hspace{1em}
\\
\textbf{\textsuperscript{3}} The University of Tokyo, Japan \hspace{1em}
\\
\texttt{\{zhong, feicheng, wan,  chu, murawaki,kuro\}@nlp.ist.i.kyoto-u.ac.jp} \\
\texttt{ying@nii.ac.jp} \\
\texttt{jiangjf@is.s.u-tokyo.ac.jp} \\ 
}

\begin{document}
\maketitle
\begin{abstract}

In this study, we investigate whether
non-English-centric LLMs
, despite their strong performance, `think' in their respective dominant language: more precisely, `think’ refers to how the representations of intermediate layers, when un-embedded into the vocabulary space, exhibit higher probabilities for certain dominant languages during generation. We term such languages as internal 
 \textbf{latent languages}.
We examine the latent language of three typical categories of models for Japanese processing: Llama2, an English-centric model; Swallow, an English-centric model with continued pre-training in Japanese; and LLM-jp, a model pre-trained on balanced English and Japanese corpora.
Our empirical findings reveal that, unlike Llama2 which relies exclusively on English as the internal latent language, Japanese-specific Swallow and LLM-jp employ both Japanese and English, exhibiting dual internal latent languages. For any given target language, the model preferentially activates the latent language most closely related to it.
In addition, we explore how intermediate layers respond to questions involving cultural conflicts between latent internal and target output languages. We further explore how the language identity shifts across layers while keeping consistent semantic meaning reflected in the intermediate layer representations.
This study deepens the understanding of non-English-centric large language models, highlighting the intricate dynamics of language representation within their intermediate layers.
\end{abstract}

\section{Introduction}


\begin{figure*}[ht]
    \centering
    \begin{minipage}{\textwidth}
        \centering
        \par\medskip
        \begin{minipage}{0.35\textwidth}
            \centering
            \includegraphics[width=\textwidth]{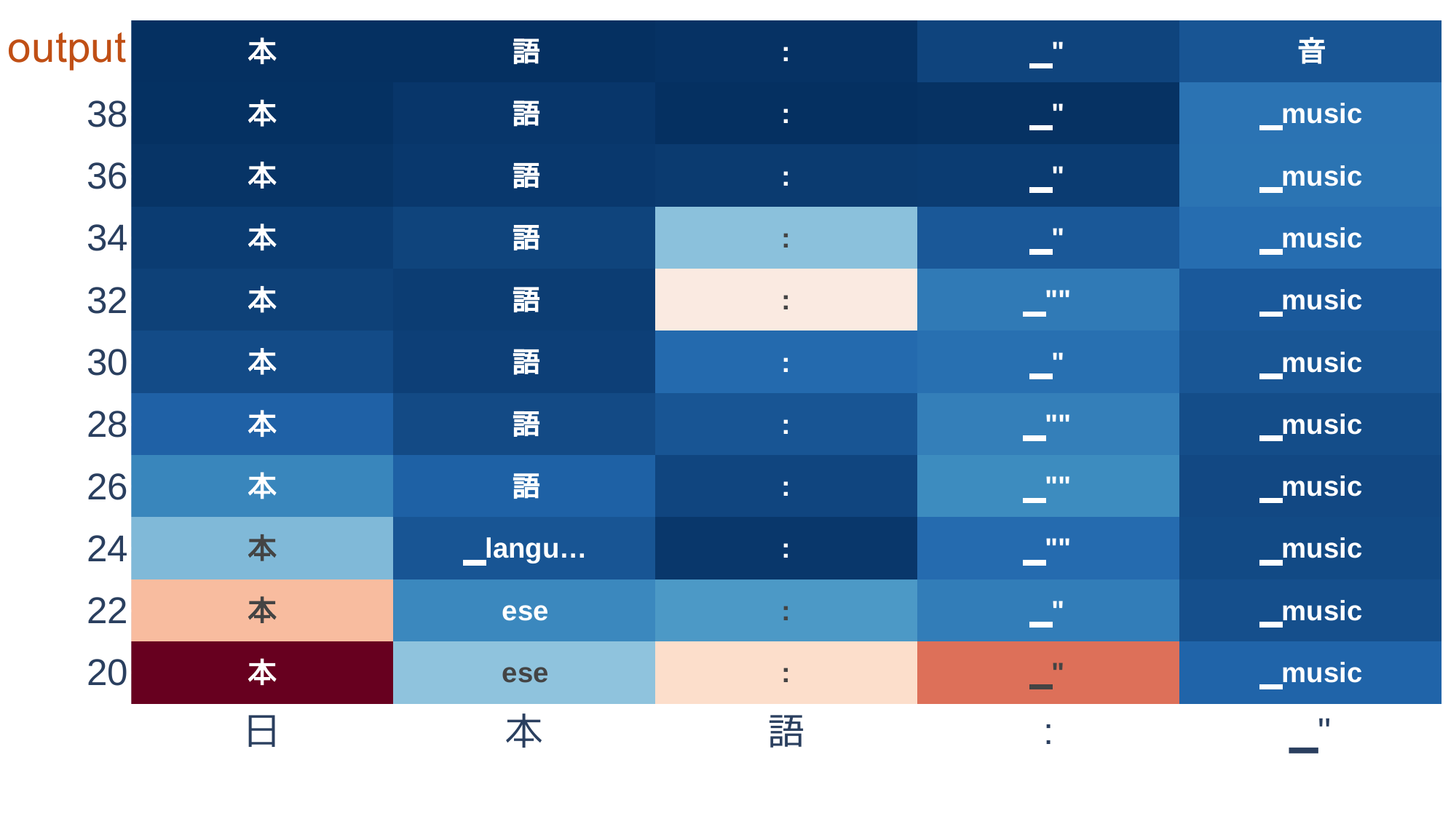}
             \par (a) Llama-2
        \end{minipage}
        \hfill
        \begin{minipage}{0.3\textwidth}
            \centering
            \includegraphics[width=\textwidth]{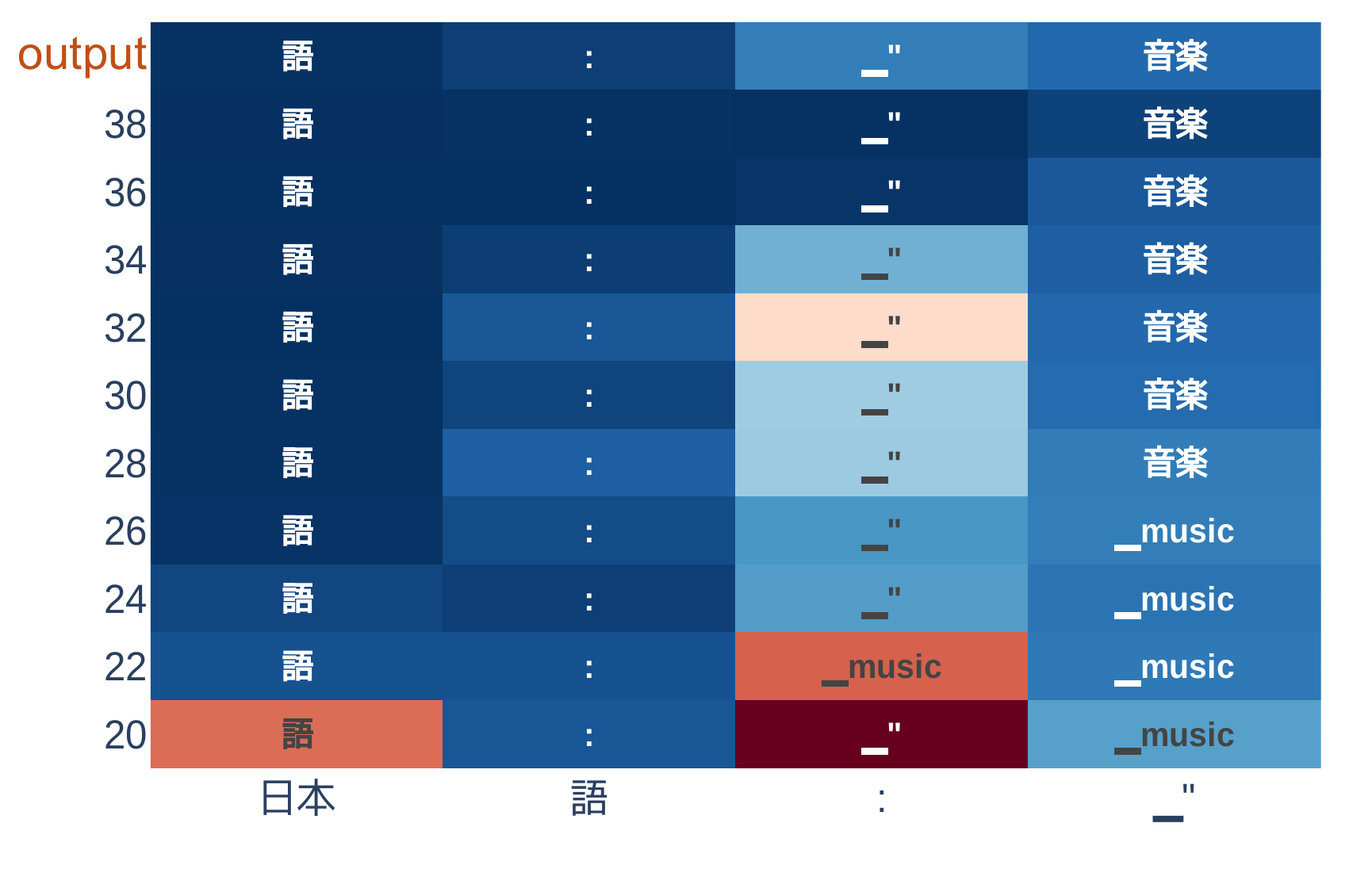}
             \par (b) Swallow
        \end{minipage}
        \hfill
        \begin{minipage}{0.3\textwidth}
            \centering
            \includegraphics[width=\textwidth]{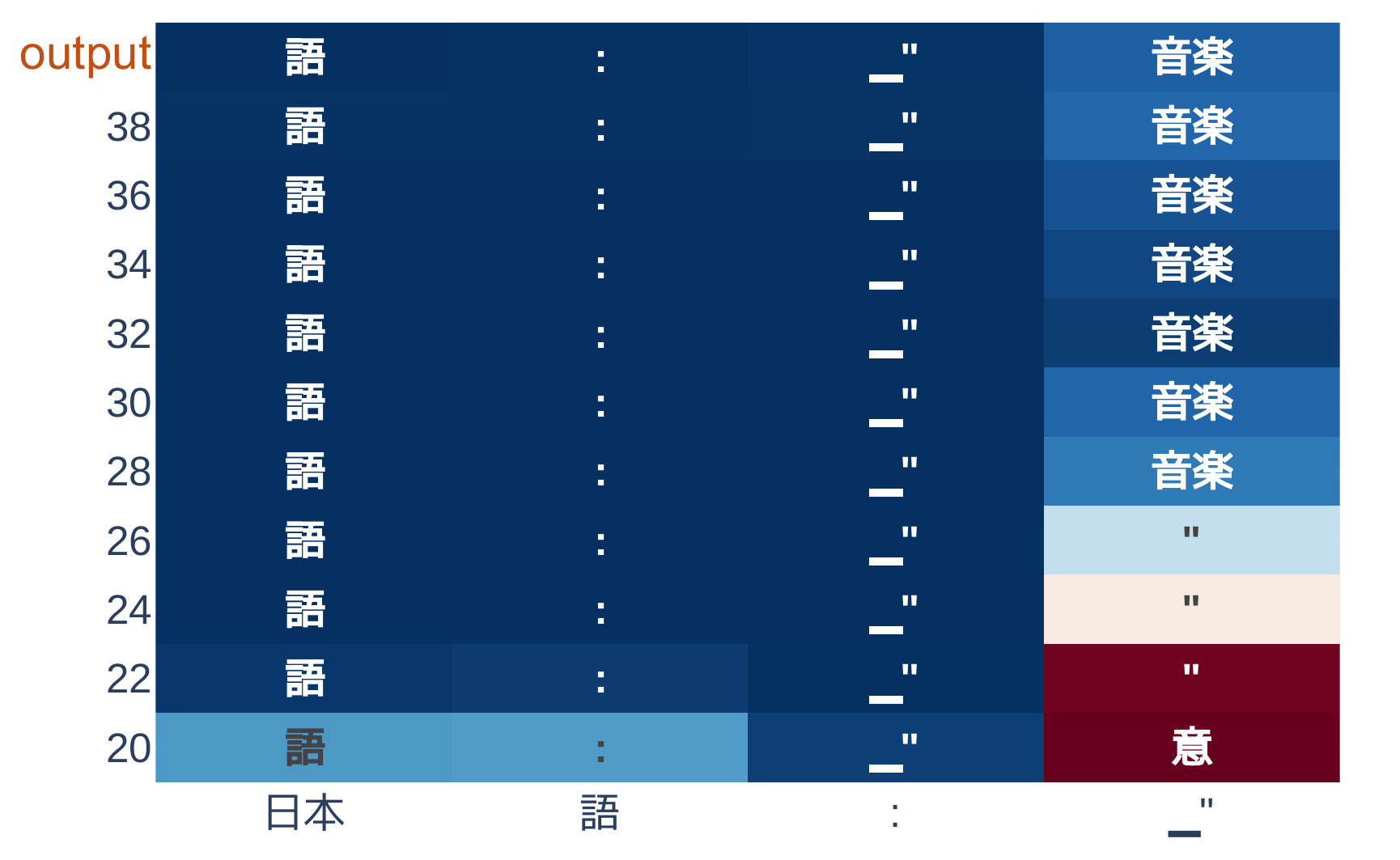}
            \par (c) LLM-jp-v2.0
        \end{minipage}
    \end{minipage}
    \caption{\textbf{Logit lens results of intermediate layers of three models,} (a) Llama-2, (b) Swallow, (c) LLM-jp. The input prompt is "Français: 'musique' - \begin{CJK}{UTF8}{min}日本語\end{CJK}: ", which is a French-to-Japanese translation task with the answer being "\begin{CJK}{UTF8}{min}音楽\end{CJK}" (music). The figure shows the highest probability token from the intermediate layers starting from layer 20. 
}
    \label{fig:combined4}
\end{figure*}

Large language models have become the prevailing approach for building NLP systems, most of which have been primarily developed for the English language. Due to the performance decline of English-centric models on non-English languages and their cultural bias towards English, researchers have increasingly focused on developing models with non-English-dominant corpora. Models that undergo continual pre-training (CPT)~\cite{sunlamol,brown2020language,csaki2024sambalingo,Chinese-LLaMA-Alpaca,hunter2023claire} or are pre-trained from scratch using non-English-dominant corpora~\cite{sengupta2023jais,yang2024qwen2, faysse2024croissantllm} often exhibit superior performance in their respective languages. 

Recent studies have investigated the underlying causes of performance decline of English-centric models on non-English languages, which show that when English-centric models process tasks of underrepresented languages such as Japanese, their intermediate layers, when un-embedded into vocabulary space, exhibit distinct patterns where the language distribution heavily skews towards English~\cite{wendler2024llamas}. This phenomenon, which we termed as the internal latent language, raises the question: \textbf{in what internal latent language do 
non-English-centric
models `think`? }
Specifically, we would like to investigate whether these models utilize the dominant language from their training corpora in their intermediate layers when processing tasks. We conduct a case study on Japanese models, chosen due to their relatively rich open-source ecosystem and the availability of training corpora information. We examine three typical categories of models that are used to
process Japanese: Llama-2~\cite{touvron2023llama}, an English-centric model; along with two Japanese-specific models Swallow~\cite{fujii2024continual}, an English-centric model with continued pre-training in Japanese; and LLM-jp~\cite{aizawa2024llm}, a model pre-trained on balanced corpora of English and Japanese.

To investigate what the LLMs `think` after each layer of transformation in the intermediate layers, we employed the logit lens method~\cite{Nostalgebraist}, which un-embeds each layer's latent representation into the vocabulary space. Given that Japanese is a combination of phonographic and logographic writing systems, we designed an alternative that extends \citet{wendler2024llamas} from single-token analysis to multi-token analysis. As shown in Figure \ref{fig:combined4}, we verify the internal latent languages of three types of models when processing Japanese and observe that they exhibit distinct behaviors: While Llama-2 uses English for pivot as shown in Figure~\ref{fig:combined4} (a), 
in contrast, the Japanese-specific model Swallow exhibits a mixed pattern, utilizing both English and Japanese within its intermediate layers, as shown in Figure~\ref{fig:combined4} (b). Meanwhile, LLM-jp, as shown in Figure~\ref{fig:combined4} (c), primarily utilizes Japanese as the internal latent language, with minimal reliance on English, highlighting its divergence from the English-centric model.

As Japanese-specific models utilize Japanese, either entirely or partially, as the latent language for processing Japanese, it is crucial to consider that both Swallow and LLM-jp have a significant proportion of English data in the pre-training corpus. An intuitive question that arises is: \textbf{which language would be their latent language 
when generating languages other than the dominant Japanese and English?
} Therefore, we introduced a setting in which non-Japanese and non-English languages such as French and Chinese, which are relatively underrepresented in the training corpora, are used as input and output languages to explore the behaviors of the internal latent language. 
Our experiments show that in intermediate layers of the model, the internal latent language of Japanese-specific models is a distribution over English and Japanese, with the probabilities of these distributions depending on their similarity to the output target language. In the final layers of the model, the internal predictions transform into the corresponding target language output.


Besides the analysis of internal latent languages in non-English-centric models, we further investigate whether certain internal latent language could cause the intermediate layers to exhibit cultural bias against the target language.
Our investigation focuses on how the model generates answers to questions where the internal latent language and the target output language conflict culturally. We observe that when the model is asked culturally related questions, the intermediate layers initially produce responses that are biased by the culture of the internal latent language. As the information moves through subsequent layers, the responses gradually align with the context of the target language. 
Moreover, given that internal latent language representations are transformed into target language representations of the same semantic meaning, we aimed to explore whether dimensions in the representation space can separate semantic and language identities. We discovered that the distribution transition within the latent layers occurs solely in sparse dimensions, while these specific dimensions are highly relevant to language identities.

In summary, the contribution of this study is listed as follows:

\begin{enumerate}
    \item We conduct a case study on Japanese-specific LLMs Swallow and LLM-jp, and confirm that they use Japanese as their internal latent language when processing Japanese.
    \item We investigate the behaviour under languages that are underrepresented in pre-training corpora, and revealed that both Swallow and LLM-jp exhibit two internal latent languages. For a specific target language, the model utilizes the latent language that is more closely related.
    \item We observe that the shift from internal latent language to target language affects the semantics of the intermediate layers of the model, while only sparse dimensions relevant to language identities undergo changes.
\end{enumerate}

\section{Related work}
\subsection{Multilingual Large Language Models}
Current frontier large language models, such as GPT-4 \cite{achiam2023gpt}, Gemini \cite{team2023gemini}, and Llama-2 \cite{touvron2023llama}, are primarily trained with English-centric corpora, with other languages constituting only a small portion of the training data. Researchers have sought to enhance these models' multilingual capabilities through various methods. One approach involves continued pre-training with second-language data~\cite{sunlamol,brown2020language,csaki2024sambalingo,Chinese-LLaMA-Alpaca,hunter2023claire}, as demonstrated by models like Swallow~\cite{fujii2024continual} based on Llama-2. Another strategy is training with bilingual data from the outset~\cite{sengupta2023jais,yang2024qwen2, faysse2024croissantllm}, exemplified by models such as LLM-jp \cite{aizawa2024llm}. Additionally, methods such as training with parallel corpora~\cite{Tower}, and expanding vocabulary followed by relearning embeddings during second-language training have been employed~\cite{extendembed}. While these approaches have proven effective, ongoing research aims to discover more efficient techniques to further improve the multilingual capabilities of large language models.

\subsection{Mechanistic Interpretability}
Mechanistic interpretability is the study of understanding how machine learning models work by analyzing their internal components and processes to elucidate the mechanisms that give rise to their behavior and predictions, encompassing research lines like superposition~\cite{elhage2022toy}, sparse autoencoders~\cite{huben2023sparse}, circuit analysis~\cite{wanginterpretability} and so on. Within these studies, logits lens \cite{Nostalgebraist} and tuned lens \cite{belrose2023eliciting} focus on decoding the probability distribution over the vocabulary from intermediate vectors of the model, aiding in the comprehension of how the model generates text in the target language.
\citet{wendler2024llamas} showed that Llama-2 models have an abstract
“concept space” that lies closer to English than to other languages. When Llama models perform tasks such as translation between non-English languages, the probabilities in the intermediate layers initially focus on the English version of the answer and gradually shift to the target language.

In this work, we expanded previous work and utilized these tools to study the distribution of latent languages in different categories of Japanese-related LLMs and examined how the probability of internal latent languages is associated with the target language. 

\begin{figure}[t!]
    \centering
    \includegraphics[width=1.0\linewidth]{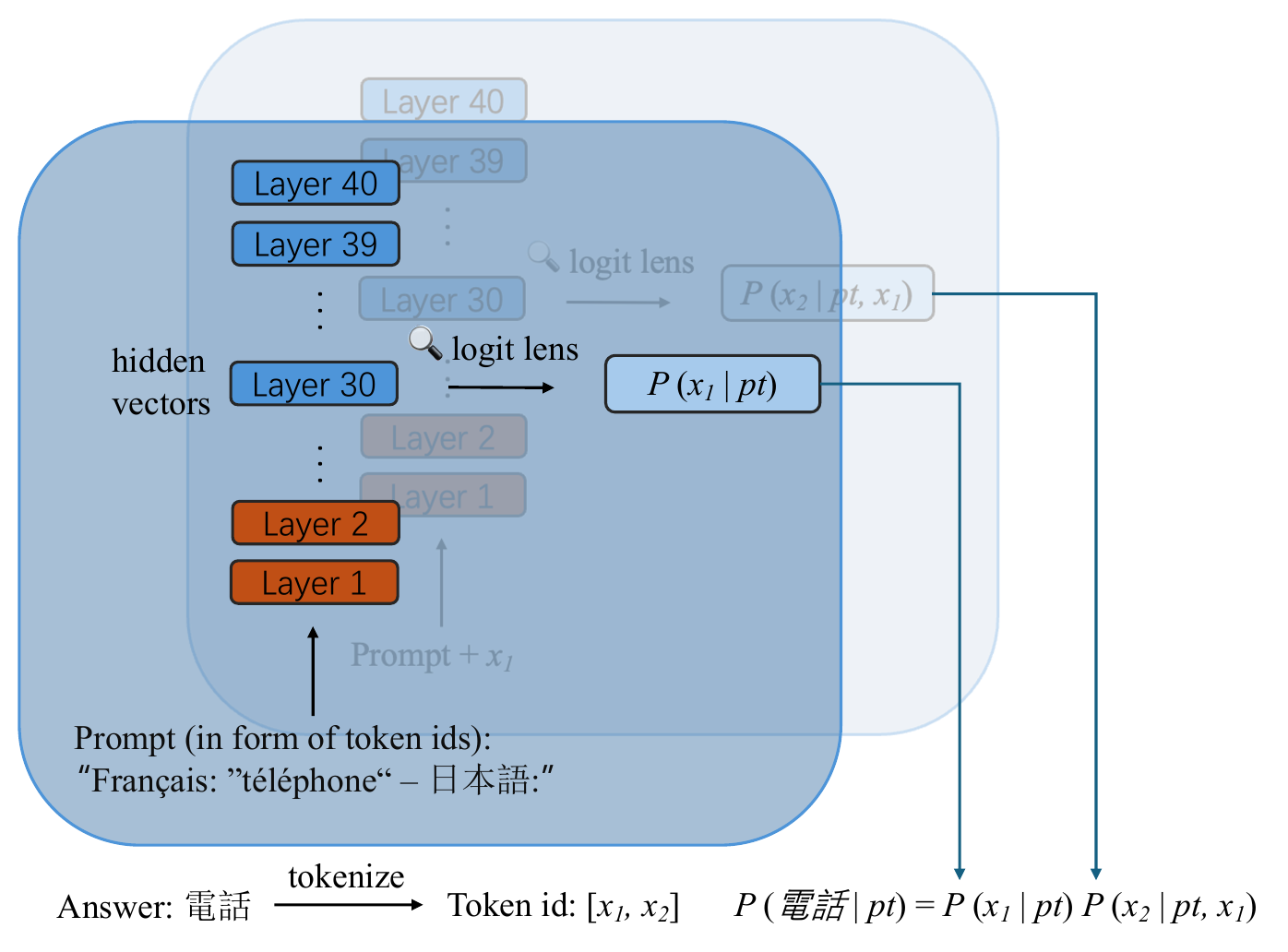}
    \caption{Example for calculating multi-token probability in intermediate layers}
    \label{fig:enter-label}
\end{figure}

\section{Method}
\subsection{Overview}
To determine which language is used in the intermediate layers of models with multiple pivot languages, we first select three types of models: (1) English-dominated model; (2) models based on an English-dominated model CPT on a second language; (3) models pre-trained from scratch with non-English-dominated corpora. We then constructed a multilingual dataset based on the models' pivot language and the degree of similarity to the primary language. The models were tested with the dataset, and the results are presented in Figure \ref{fig: translation}, \ref{fig: repitition}, and \ref{fig: cloze}.

\subsection{Logit Lens}

To convert vectors into tokens, the model’s output layer uses an unembedding matrix to project the hidden vectors, which are propagated within the model, onto the dimensions of the vocabulary. Then, softmax is applied to calculate the probabilities and generate the output token. And this is called unembedding.
Since the hidden vectors passed between the intermediate layers of the model have the same dimensions as the output vectors. By applying the same unembedding operation to these hidden vectors, we can obtain some information about what happen in the intermediate layers. \textit{Logit lens} is a tool designed to achieve this purpose. We use a similar method to obtain the predicted token probability distribution from the intermediate layers.

\subsubsection{Measuring Multi-token Sequence Probability}

The vocabulary of a model is limited. A single word from non-primary languages often requires multiple tokens for representation. Besides, many Chinese and Japanese characters share the same form. Additionally, the meaning of a single character in Chinese and Japanese is always not clearly defined. So we use two-character phrases to ensure a more precise expression. Based on the above reasons, single-token level probability calculation does not meet our requirements. Consequently, we designed a method to calculate the generation probability of a token sequence in the intermediate layers.

The method begins by using the model's tokenizer to decompose a word or phrase into a sequence of token IDs. Given a prompt, for a token ID sequence \([ \textit{x}_1, \textit{x}_2, \ldots, \textit{x}_n ]\), the probability \(\textit{p}_1\) of token \(\textit{x}_1\) is first obtained at layer \(\textit{i}\) using the logit lens method on the hidden vectors. Subsequently, token \(\textit{x}_1\) is input into the model as the predicted token, and the probability \(\textit{p}_2\) of token \(\textit{x}_2\) is calculated at layer \(\textit{i}\). This process is repeated iteratively. The final probability of generating the token sequence \([ \textit{x}_1, \textit{x}_2, \ldots, \textit{x}_n ]\) at layer \(\textit{i}\) is then determined as the product of individual probabilities, \(\textit{p}_1 \times \textit{p}_2 \times \cdots \times \textit{p}_n\).

\subsection{Categorization of Multilingual Large Language Models}

Based on their training corpora and construct method, we classify language models into three types:

\vspace{0.5em} 
\noindent\textbf{English-Centric Models.} These models, such as Llama2, the majority of their training data is in English, making them highly proficient in generating and understanding English text.

\vspace{0.5em} 
\noindent\textbf{Multilingual CPT Models.} These models are built upon an English-Centric Model and undergo continued pre-training on a second language or more to enhance multilingual ability.

\vspace{0.5em} 
\noindent\textbf{Balanced Multilingual Models.} These models are trained on a roughly equal amount of tokens from two or more languages, aiming to achieve balanced proficiency across these languages.

This categorization is based on different training corpora configurations, so we can study how the training corpora influence the latent language probabilities and overall performance of language models on multilingual tasks.

\begin{table*}
\centering
\begin{tabular}{|l|l|r|r|r|r|c|}
\hline
\textbf{Model Category} & \textbf{Model} & \multicolumn{3}{c|}{\textbf{Proportion in pre-training data}} & \textbf{\#Token} & \textbf{From scratch} \\ \cline{3-5}
               &          & \textbf{En} & \textbf{Ja} & \textbf{Other} &          &          \\ \hline
English-centric & Llama 2  & 89.70\%  & 0.10\%  & 10.20\%  & 2,000B & Yes      \\ \hline
Multilingual CPT & Swallow  & 10\%     & 90\%     & 0\%      & 100B   & Llama-2 based       \\ \hline
Balanced Multilingual & LLM-jp & 50\%     & 50\%     & 0\%      & 300B   & Yes      \\ \hline
\end{tabular}
\caption{Categorization of multilingual models based on language proportion and training strategy}
\label{tab:multilingual_models}
\end{table*}

\begin{figure*}[t]
    \centering
    \begin{minipage}{\textwidth}
        \centering
        (a) Translation: Fr -> Ja
        \par\medskip
        \begin{minipage}{0.3\textwidth}
            \centering
            \includegraphics[width=\textwidth]{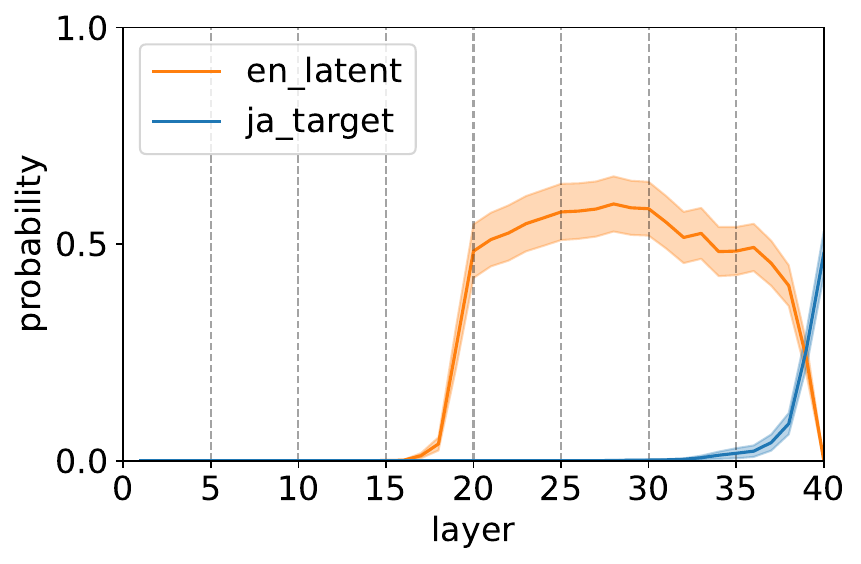}
        \end{minipage}
        \hfill
        \begin{minipage}{0.3\textwidth}
            \centering
            \includegraphics[width=\textwidth]{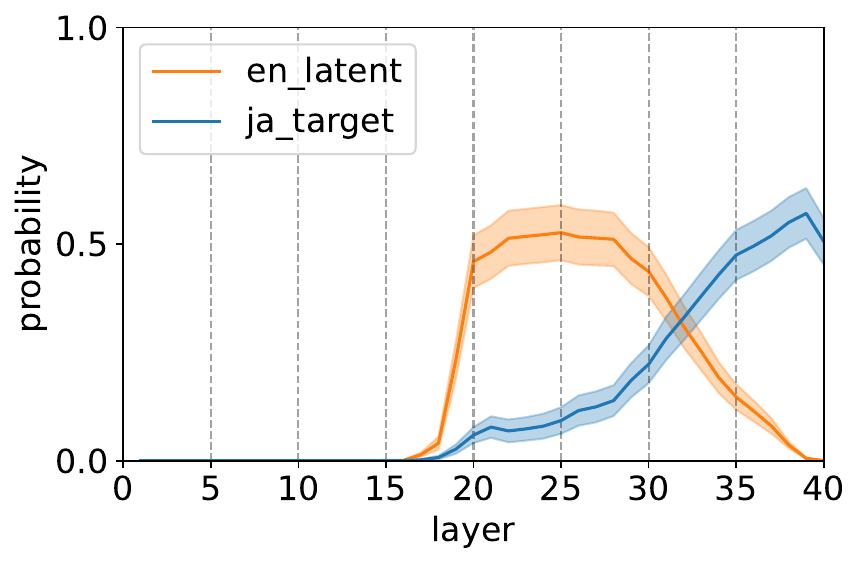}
        \end{minipage}
        \hfill
        \begin{minipage}{0.3\textwidth}
            \centering
            \includegraphics[width=\textwidth]{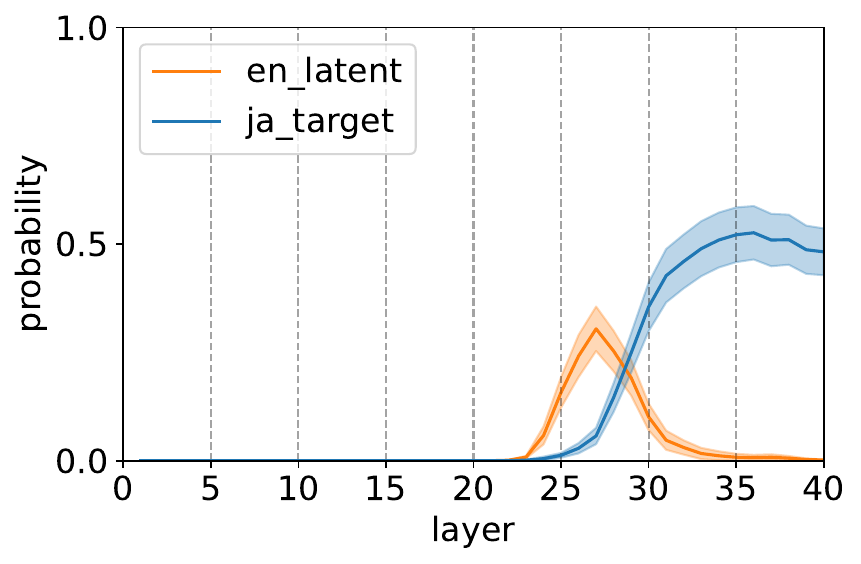}
        \end{minipage}
    \end{minipage}

    \begin{minipage}{\textwidth}
        \centering
        (b) Repetition: Ja
        \par\medskip
        \begin{minipage}{0.3\textwidth}
            \centering
            \includegraphics[width=\textwidth]{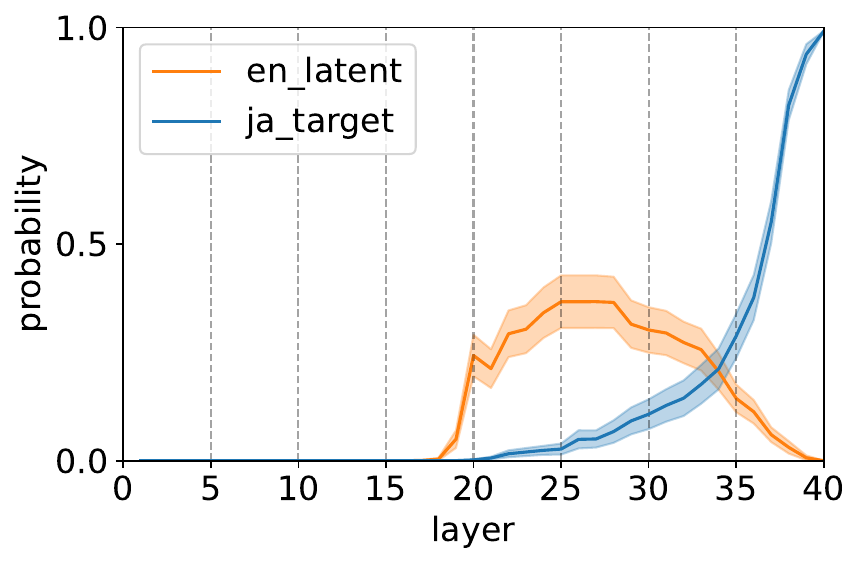}
        \end{minipage}
        \hfill
        \begin{minipage}{0.3\textwidth}
            \centering
            \includegraphics[width=\textwidth]{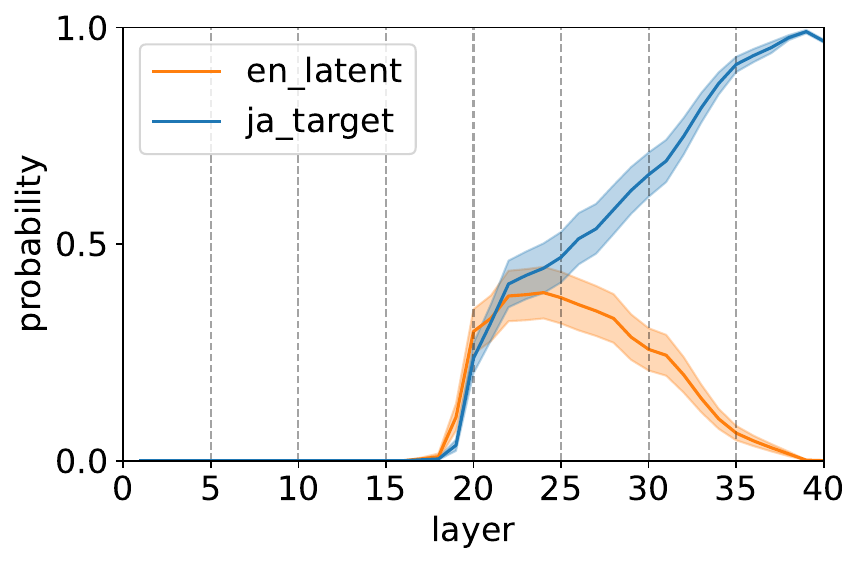}
        \end{minipage}
        \hfill
        \begin{minipage}{0.3\textwidth}
            \centering
            \includegraphics[width=\textwidth]{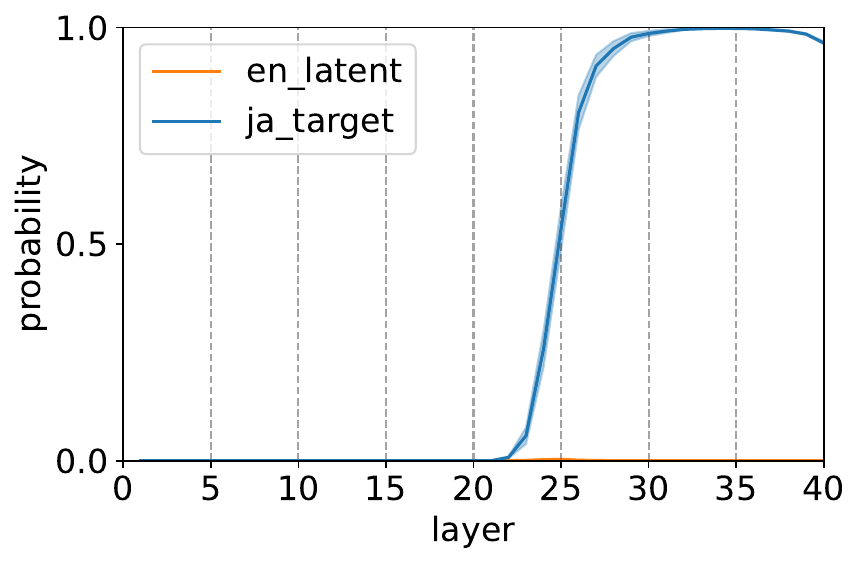}
        \end{minipage}
    \end{minipage}

    \begin{minipage}{\textwidth}
        \centering
        (b) Cloze: Ja
        \par\medskip
        \begin{minipage}{0.3\textwidth}
            \centering
            \includegraphics[width=\textwidth]{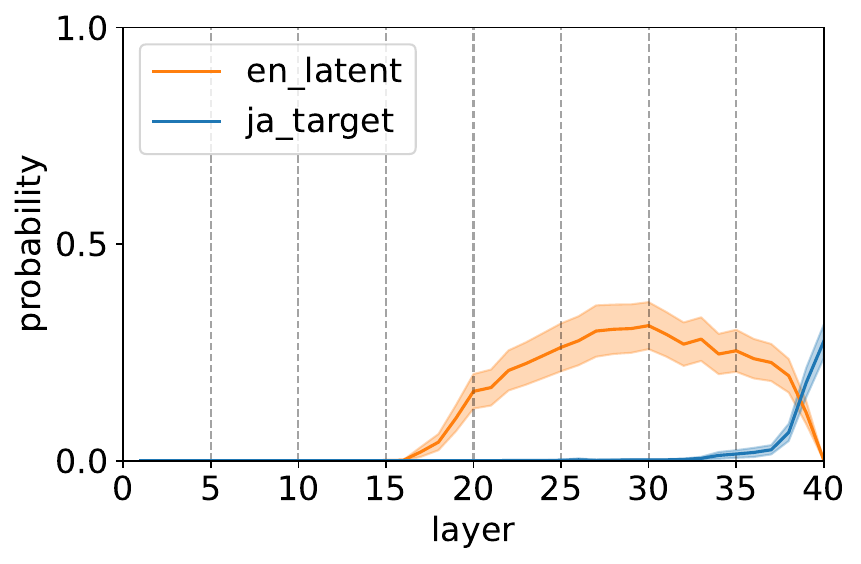}
            \par Llama-2-13b
        \end{minipage}
        \hfill
        \begin{minipage}{0.3\textwidth}
            \centering
            \includegraphics[width=\textwidth]{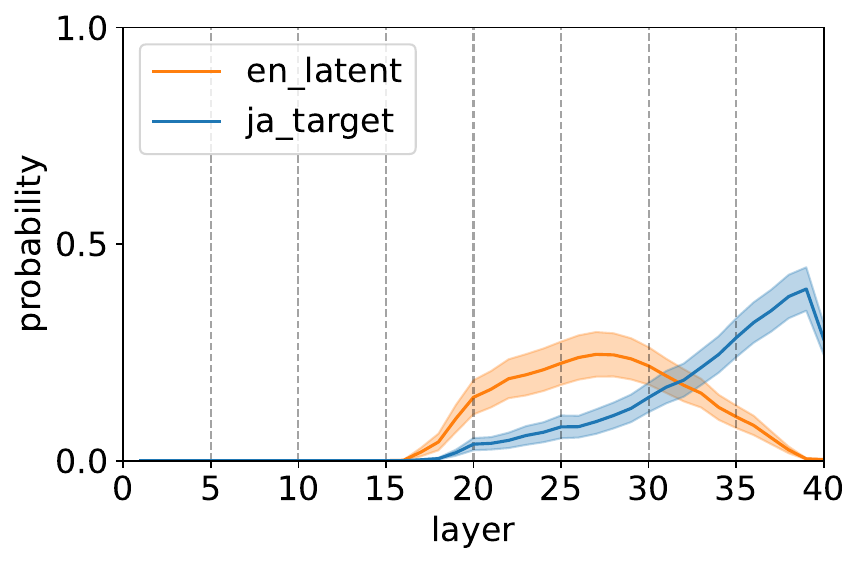}
            \par Swallow-13b
        \end{minipage}
        \hfill
        \begin{minipage}{0.3\textwidth}
            \centering
            \includegraphics[width=\textwidth]{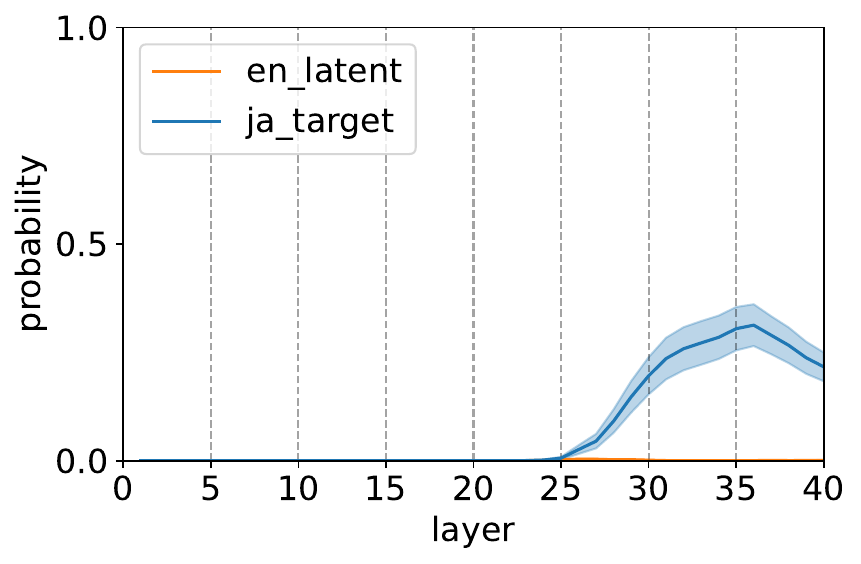}
            \par LLM-jp-v2.0
        \end{minipage}
    \end{minipage}
    \caption{\textbf{Comparison of English-centric and non-English-centric models when processing Japanese,} (a) French to Japanese translation, (b) Japanese repetition, (c) Japanese cloze task. X-axes stand for layer index and y-axes stand for probability of answer in each language. Error bars show 95\% Gaussian confidence intervals over totally 166 input examples.}
    \label{fig: case of jp}
\end{figure*}

\subsection{Dataset Construction}
We aim to study which language is used in the intermediate layers when non-English-dominated models process different languages. Naturally, the constructed dataset should first include the model's pivot language itself.
The pivot languages of the training data for the models we choose are English and Japanese, so these two languages must be considered. For each language, we select a similar one to investigate whether the target language affects the probabilities of these two languages in the intermediate layers. Specifically, we choose French as the similar language for English and Chinese for Japanese. Because Chinese and Japanese share common characters, we first prepared a set of non-overlapping Chinese-Japanese word pairs that have the same meaning but different characters. We construct this based on \textit{Database of Japanese Kanji Vocabulary in Contrast to Chinese} (JKVC) \begin{CJK}{UTF8}{min}\cite{JKVC2020}\end{CJK}. Then, we use GPT-4 to do translation and obtain the corresponding English and French words or phrases, and check if they are correct. Consequently, we obtain the parallel data like in the following frame.
\begin{framed}
\noindent
Français: "principe"\\
English: "principle"\\
\begin{CJK}{UTF8}{min}日本語: "原則"\end{CJK}\\
\begin{CJK}{UTF8}{gbsn}中文: "原则"\end{CJK}
\end{framed}

\vspace{0.5em} 
\noindent\textbf{Prompt design.} We examine the models on three tasks and with the following prompt format, following previous studies~\cite{wendler2024llamas}. We demonstrate the following three tasks, and the corresponding answers for three examples will be the same Japanese word \begin{CJK}{UTF8}{min}"原則"\end{CJK} (principle).

\vspace{0.5em} 
\noindent\textit{Translation task:}
We use four-shot prompt in this task. The few-shot format can make it easier for the model to understand the required task without adding additional instructions in other languages, minimizing unnecessary interference. 

When constructing prompts, we use a hyphen to connect the input language line and the target language line to form a one-shot. In the fifth shot, we omit the answer after the last colon, leaving it for the model to predict.
For example: 
\begin{framed}
\noindent
Français: "principe" - \begin{CJK}{UTF8}{min}日本語:"\end{CJK}
\end{framed}



\vspace{0.5em} 
\noindent\textit{Repetition task:}
Four-shot prompt for repetition is similar to the translation one but repeat the same language twice. 
In the last shot, we omit the answer after the last colon, leaving it for the model to predict.
For example:
\begin{framed}
\noindent\\\textit{\begin{CJK}{UTF8}{min}日本語: "原則" - 日本語:"\end{CJK}}
\end{framed}

\vspace{0.5em} 
\noindent\textit{Cloze task:}
For the Cloze task, we ask GPT-4 to generate a description for each word in each language. Each described word is placed at the beginning of the description. We then mask the word in the description and ask the models to generate the target word. This task is similar to a QA task, which requires common knowledge. To maintain consistency with previous work, we use two-shot prompting in this task.
In the last shot, we omit the answer after the last colon, leaving it for the model to predict. For example:
\begin{framed}
\noindent\\\textit{\begin{CJK}{UTF8}{min}"\_\_"は、基本的なルールや信念です。答え: "\end{CJK}}
\end{framed}

\section{Experiment Settings}
\noindent\textbf{Details of the Models.} We selected one model from each of the three types mentioned earlier, Llama-2, Swallow, and LLM-jp-v2.0. We use the 13B size of all three models consistently for fair comparison. All models have 40 layers and a word embedding dimension of 5120. Llama-2 has a vocabulary size of 32,000 tokens. 43,176 tokens for Swallow and 96,867 tokens for LLM-jp-v2.0. And 8-bit quantization \cite{8bit} is used in our experiments. Other details are shown in Table ~\ref{tab:multilingual_models}.

\noindent\textbf{Details of Dataset.}
The dataset contains parallel phrases in four languages—English, French, Japanese, and Chinese—along with their corresponding descriptions. It is used to consisting prompts for translation, repetition, and cloze tasks. The total dataset size is 166.

\section{Results}
\noindent\textbf{Main experiment 1 on Specific Dominant Language.}
To investigate which internal latent language is used when processing Japanese, we conduct experiments on: Translation task with French as input language and Japanese as output target language; Repetition task and cloze task with Japanese input and Japanese target output.

\noindent\textbf{Main Experiment 2 on non-Dominant Languages.}
To investigate which internal latent language is used when processing non-dominant languages in the corpora, we conduct experiments on: Translation task in two directions, French as input language with Chinese as output target language, and vice versa; Repetition task and cloze task with monolingual input and target output, on French and Chinese separately.


\subsection{Main Experiment 1: Analysis on Specific Dominant Language -- Japanese}
\label{sec: main results}
 As shown in Figure \ref{fig: case of jp}, we compare the internal latent language behaviors of English-centric Llama and Japanese-specific models when processing all of the three tasks (translation, repetition and cloze) with Japanese set as the target language. Llama, which is an English-dominant model, exhibits using English as pivot in its intermediate layers. In contrast, Swallow, which underwent CPT in Japanese, demonstrates a noticeable probability of Japanese in its intermediate layers. For LLM-jp, which is trained on bilingual English-Japanese data, English probabilities are nearly absent in the intermediate layers during monolingual repetition and cloze tasks, and Japanese dominates the intermediate layer distribution. This indicates that these Japanese-specific models lean to utilize Japanese more as the latent language when processing Japanese, exhibiting unique characteristics compared to English-centric models

\begin{figure*}[ht]
    \centering
    \begin{minipage}{\textwidth}
        \centering
        (a) Cloze: Zh
        \par\medskip
        \begin{minipage}{0.3\textwidth}
            \centering
            \includegraphics[width=\textwidth]{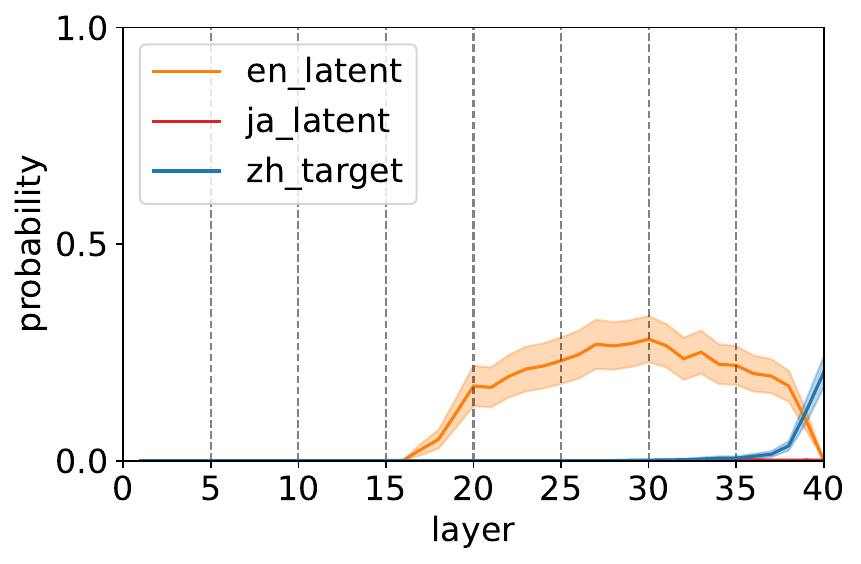}
        \end{minipage}
        \hfill
        \begin{minipage}{0.3\textwidth}
            \centering
            \includegraphics[width=\textwidth]{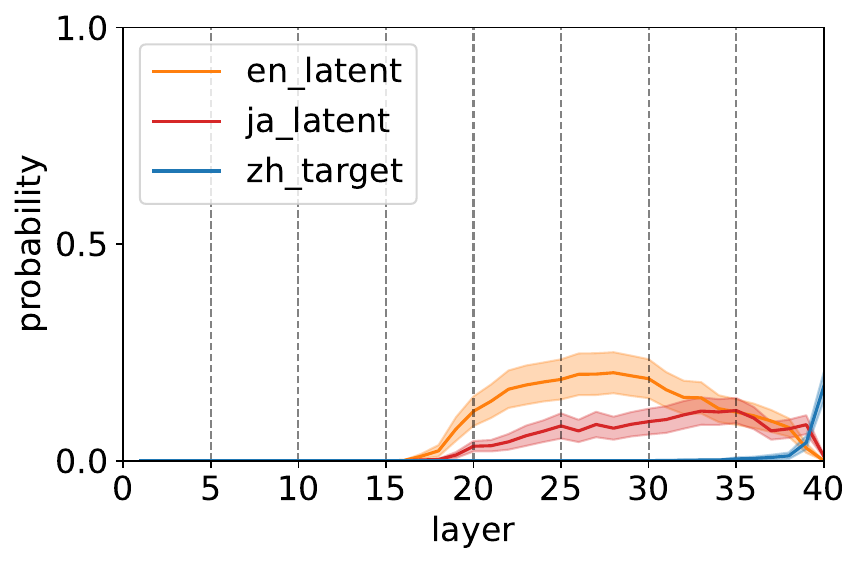}
        \end{minipage}
        \hfill
        \begin{minipage}{0.3\textwidth}
            \centering
            \includegraphics[width=\textwidth]{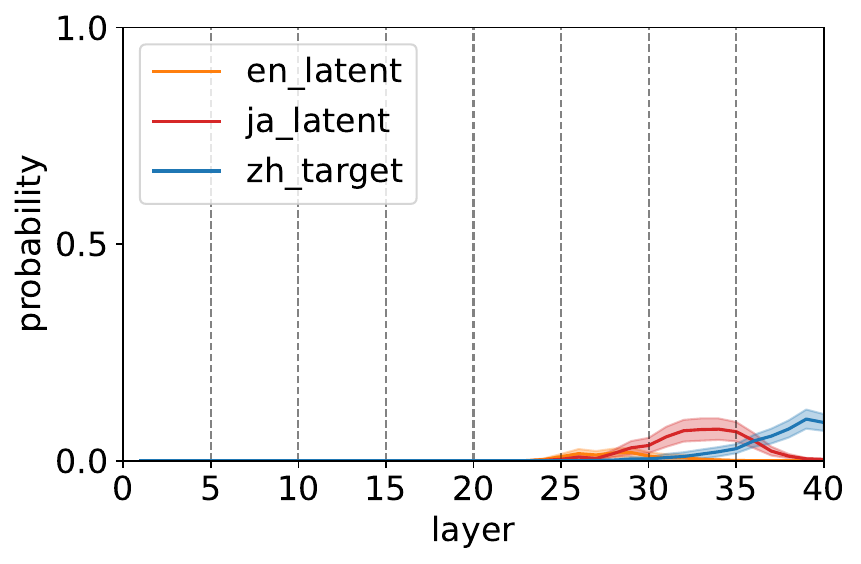}
        \end{minipage}
    \end{minipage}

    \par\bigskip

    \begin{minipage}{\textwidth}
        \centering
        (b) Cloze: Fr
        \par\medskip
        \begin{minipage}{0.3\textwidth}
            \centering
            \includegraphics[width=\textwidth]{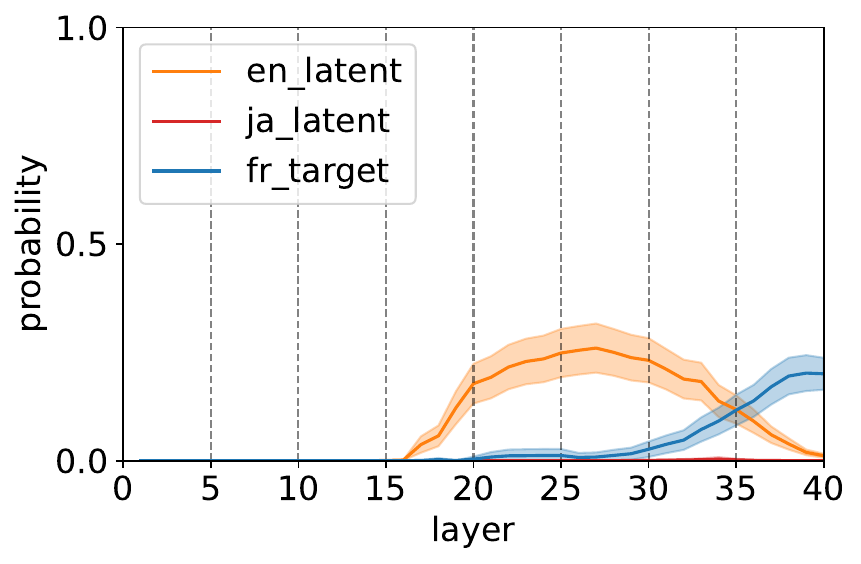}
            \par Llama-2-13b
        \end{minipage}
        \hfill
        \begin{minipage}{0.3\textwidth}
            \centering
            \includegraphics[width=\textwidth]{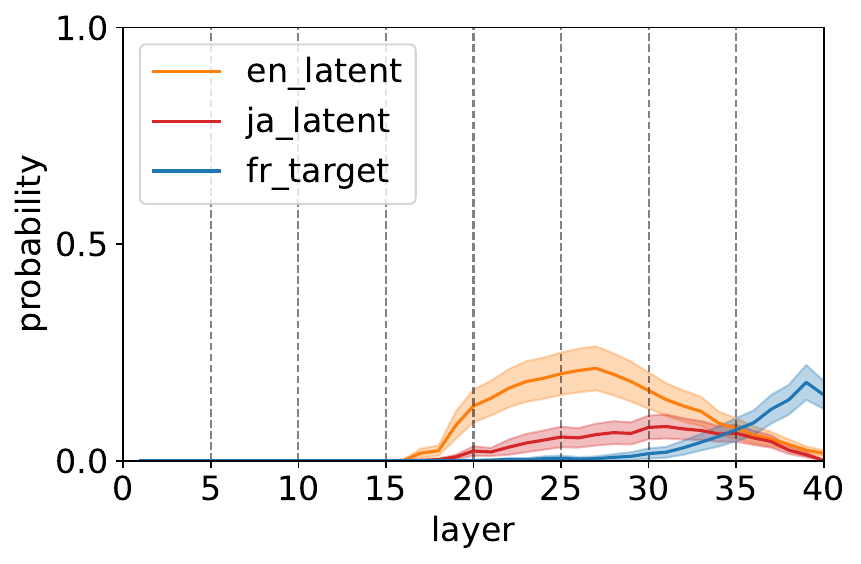}
            \par Swallow-13b
        \end{minipage}
        \hfill
        \begin{minipage}{0.3\textwidth}
            \centering
            \includegraphics[width=\textwidth]{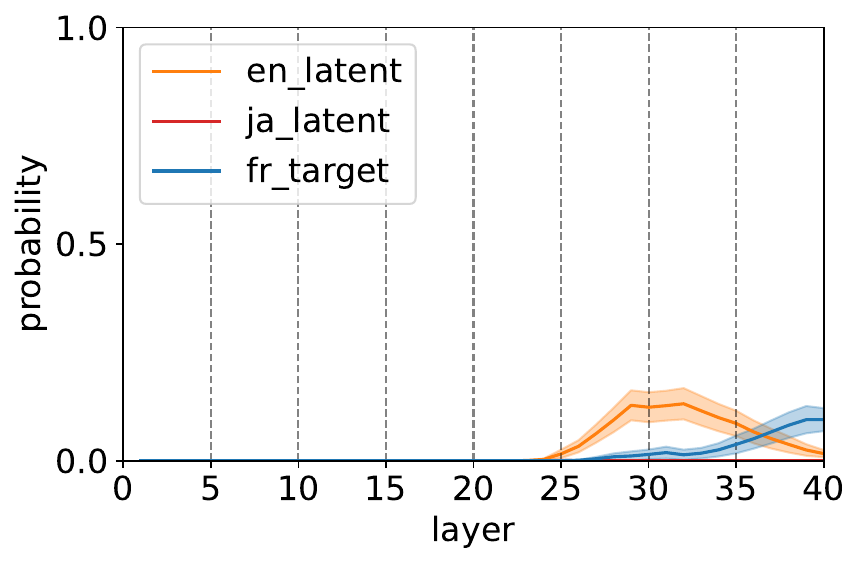}
            \par LLM-jp-v2.0
        \end{minipage}
    \end{minipage}
    \caption{\textbf{Language probabilities for three types of models in cloze task,} (a) French cloze task, (b) Chinese cloze task. X-axes stand for layer index and y-axes stand for probability of answer in each language. Error bars show 95\% Gaussian confidence intervals over totally 166 input examples.}
    \label{fig: cloze}
\end{figure*}

\begin{figure*}[ht]
    \centering
    \begin{subfigure}[t]{0.32\textwidth}
        \centering
        \includegraphics[width=\linewidth]{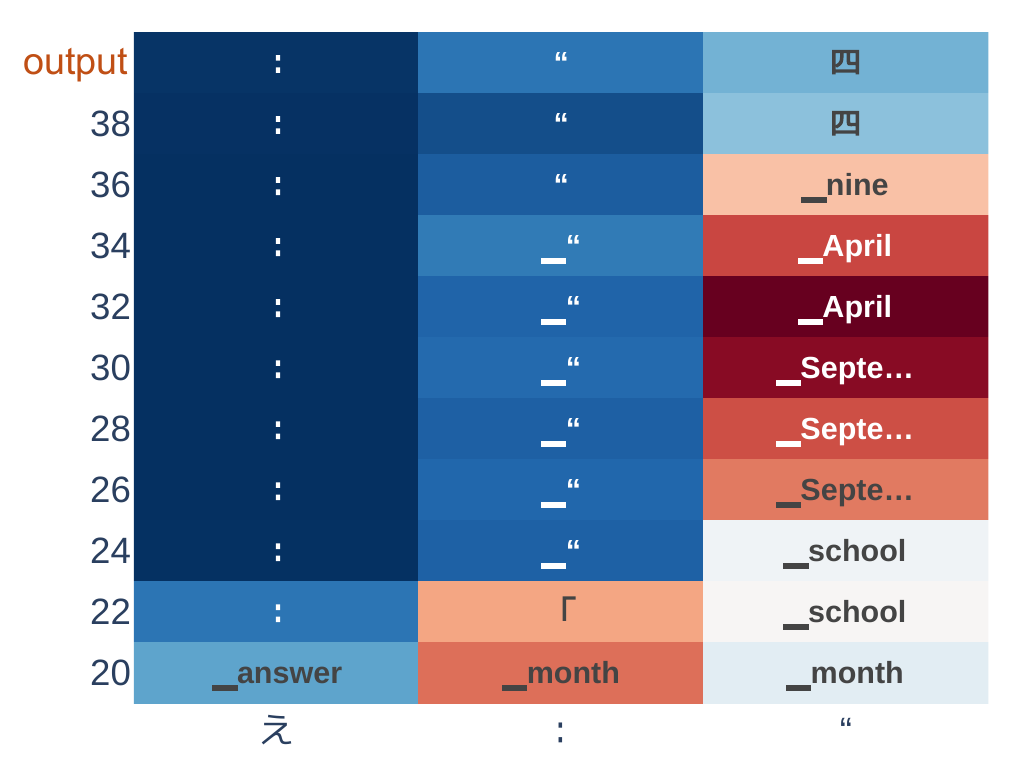}
        \caption{English-centric: Llama-2-13b}
    \end{subfigure}%
    \begin{subfigure}[t]{0.32\textwidth}
        \centering
        \includegraphics[width=\linewidth]{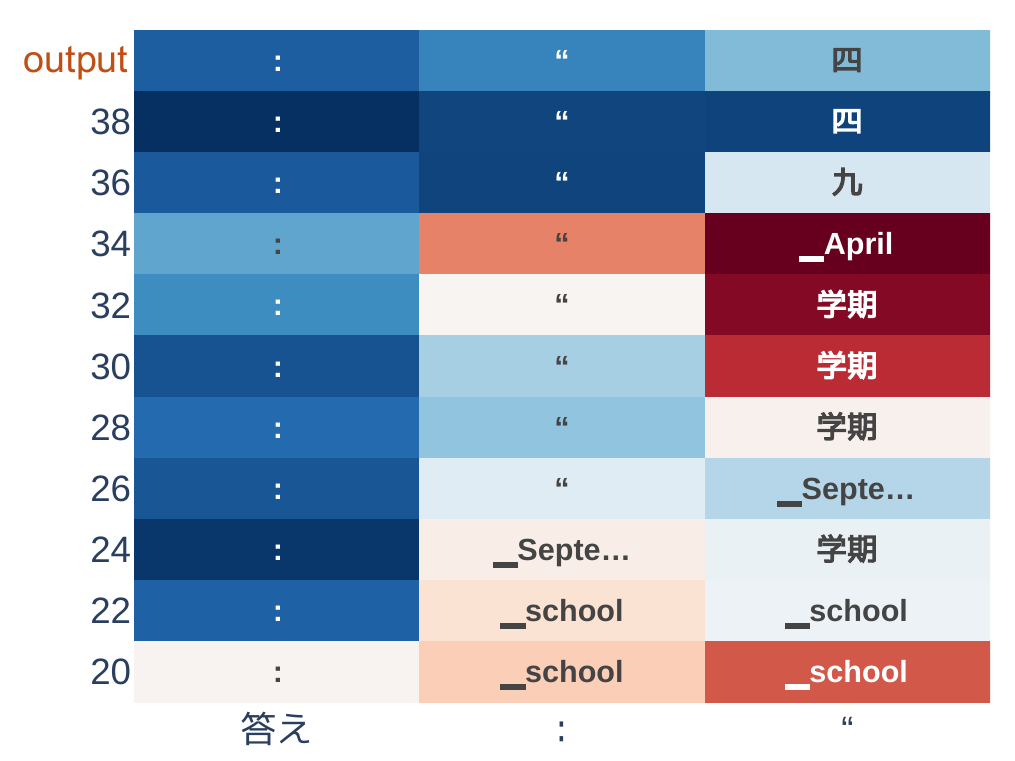}
        \caption{Multilingual CPT: Swallow-13b}
    \end{subfigure}%
    \begin{subfigure}[t]{0.32\textwidth}
        \centering
        \includegraphics[width=\linewidth]{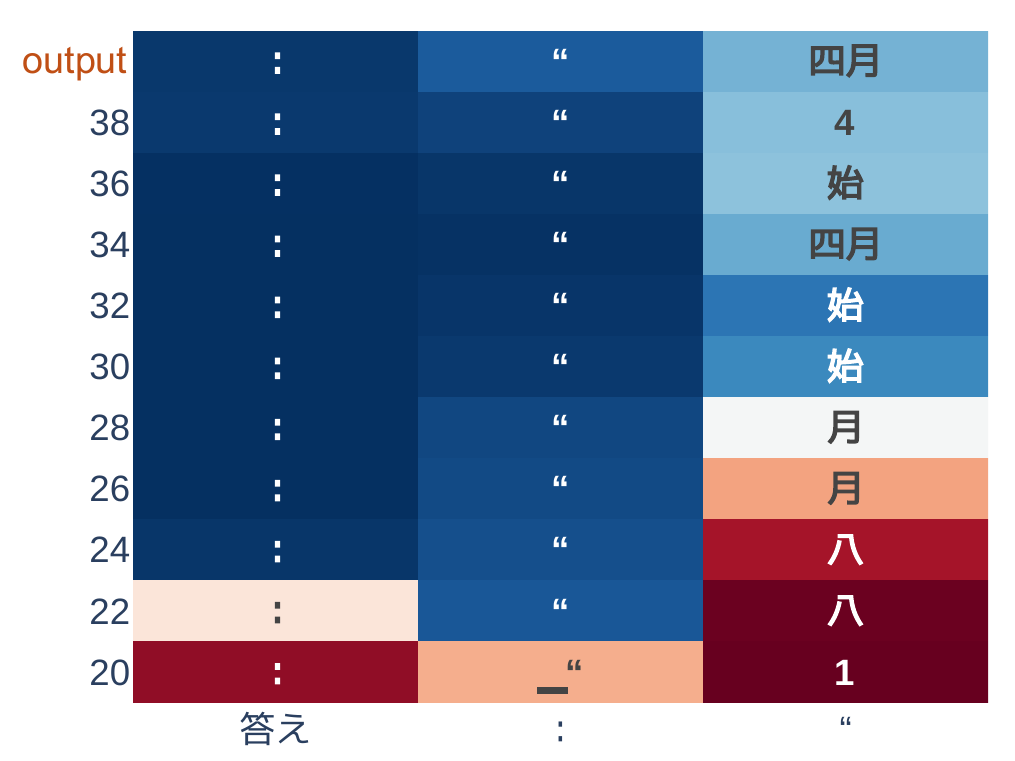}
        \caption{Balanced Multilingual: LLM-jp-v2.0}
    \end{subfigure}
    \caption{\textbf{Results of culture conflict question,} we use one shot format prompt and the question is "\begin{CJK}{UTF8}{gbsn}日本の学校新学期が始まる月は：＿月、答え：“\end{CJK}" (The month when the new school term starts in Japan is: \_ month, answer: '). The correct answer is "\begin{CJK}{UTF8}{gbsn}四\end{CJK}" (april). The colors in the figures represent entropy. Blue indicates that the probability is concentrated on the top tokens, while red means that the probability is dispersed across the vocabulary.}
    \label{fig: culture}
\end{figure*}

\subsection{Main Experiment 2: Analysis on non-Dominant Languages}

 We further investigate which internal latent language the models use when processing non-dominant languages in the corpora. 
We show translation task results in appendix Figure~\ref{fig: translation}, repetition task results in appendix Figure~\ref{fig: repitition} and cloze task results in Figure~\ref{fig: cloze}.
In all tasks of Llama-2 model, the probability of Japanese is nearly negligible and English is the internal pivot language. In contrast, the Swallow model utilizes both English and Japanese as internal pivot language in all tasks. Swallow exhibits a notable probability of Japanese in the intermediate layers, although still lower than that of English. In the LLM-jp model, which is trained on equal amounts of English and Japanese data, the probability distributions for Japanese and English in the intermediate layers are significantly influenced by the target language. Notably, when the target language is Chinese, the probability of Japanese is considerably higher than that of English;  when processing French, the probability of English is higher than that of Japanese. The model tends to utilize the internal latent language that is more closely related to the target language.

The only exception is the Swallow's Fr <-> Zh translation result shown in Figure~\ref{fig: translation}. Compared to when the target language is Chinese, Japanese probabilities in intermediate layers is higher when target language is French. This may be due to the presence of certain specific content, such as French-Chinese word pairs, mixed in Swallow's training corpus. However, this hypothesis is difficult to verify. We will conduct some additional tests on other language and test other CPT models to reach a reliable conclusion.

\begin{figure*}[hbt]
    \centering
    \includegraphics[width=1.0\linewidth]{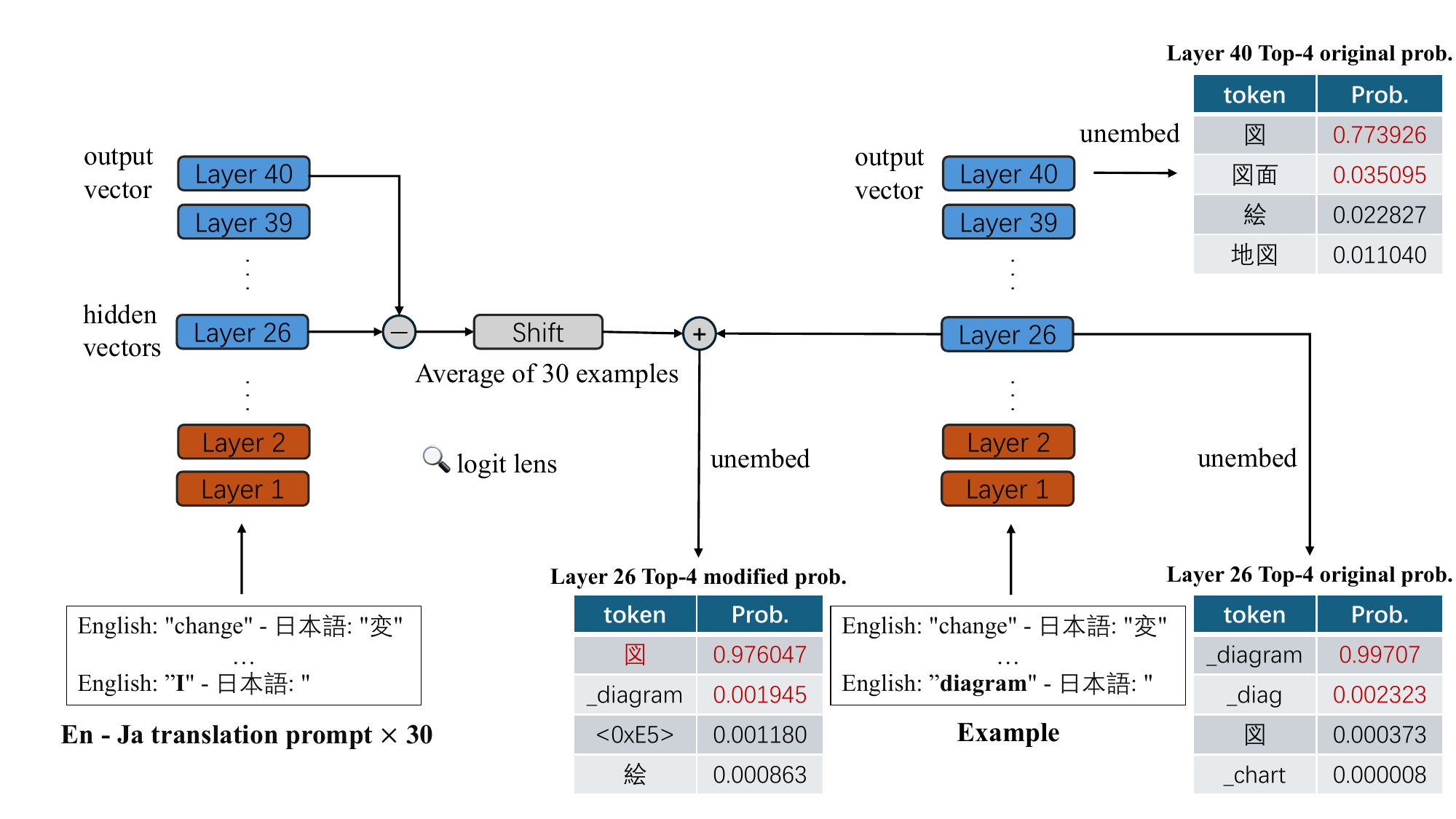}
    \caption{\textbf{Example of transiting latent probability of Swallow into the target language by adding an averaged shift vector in the translation task.}}
    \label{fig: how to shift}
\end{figure*}

\subsection{How Is Culture Conflict QA Solved?}
Since the models 'think' in pivot languages in its intermediate layers, whether this affects the model's reasoning in QA tasks is a question worth discussing. Because some questions can have different answers in different cultural contexts across languages. 
Thus, we create a small dataset of questions with different answers in different cultural contexts and use the logit lens to observe the intermediate layers of the models.

As shown in Figure \ref{fig: culture}, we ask the models about the start date of the school year in Japan with Japanese prompt. In Japan, the new school term begins in April. Even when asked about the start of the new academic year in Japan, Llama-2's English-dominant intermediate layers prefer the answer "September/nine," which is the typical start date for American schools, if you ask Llama-2 about American schools, it will answer September. The correct answer for Japan only appears in the latter layers where the probability is concentrated on the target language. In Swallow, the wrong answer \begin{CJK}{UTF8}{gbsn}"九"(nine)\end{CJK} only appear once in layer 36. In contrast, the bilingual-centric LLM-jp does not exhibit this issue. You can see in the early layers that other numbers like \begin{CJK}{UTF8}{gbsn}"八"(eight)\end{CJK} and 1 appear. But it is likely just due to the chaotic state in the early layers before the answer is determined. This indicates that, for such questions, the knowledge in the primary language context significantly influences the model's predictions. This provides an internal perspective on why operations like knowledge editing should focus on the model's primary language.

\subsection{Can Semantic and Language Identity Dimensions Be Recognized?}
As we observe in section \ref{sec: main results}, once the models obtain a latent language representation in certain middle layer, the following layers until the top (i.e., output layer) mainly involves converting the representation back to the target language. Intuitively, we are interested in investigating how this multilingual transition is exactly happened and whether it is possible to decompose the semantic and language identity dimensions.

So, we monitor the changes in hidden vectors as the probabilities shift from being concentrated in the primary latent language at intermediate layers to the target language at the output layer. We used 30 pairs of synonymous English-Japanese words. Since many Japanese Kanji characters in Llama-2 are represented in Unicode, we utilized Swallow for testing to ensure clearer results. Each of these 30 synonym pairs is represented as a single token in Swallow's vocabulary. We used the same prompts as in the translation task, input them into the model, and obtained the hidden vectors from the output layer and intermediate layers for comparison. In the previous translation task, we observed that the peak probability for latent English typically occurs around the 25th layer. We calculated the difference between the hidden vectors of the 40th layer and the 26th layer, and computed the average for these 30 samples. 

\begin{figure}[t]
    \centering
    \includegraphics[width=1.0\linewidth]{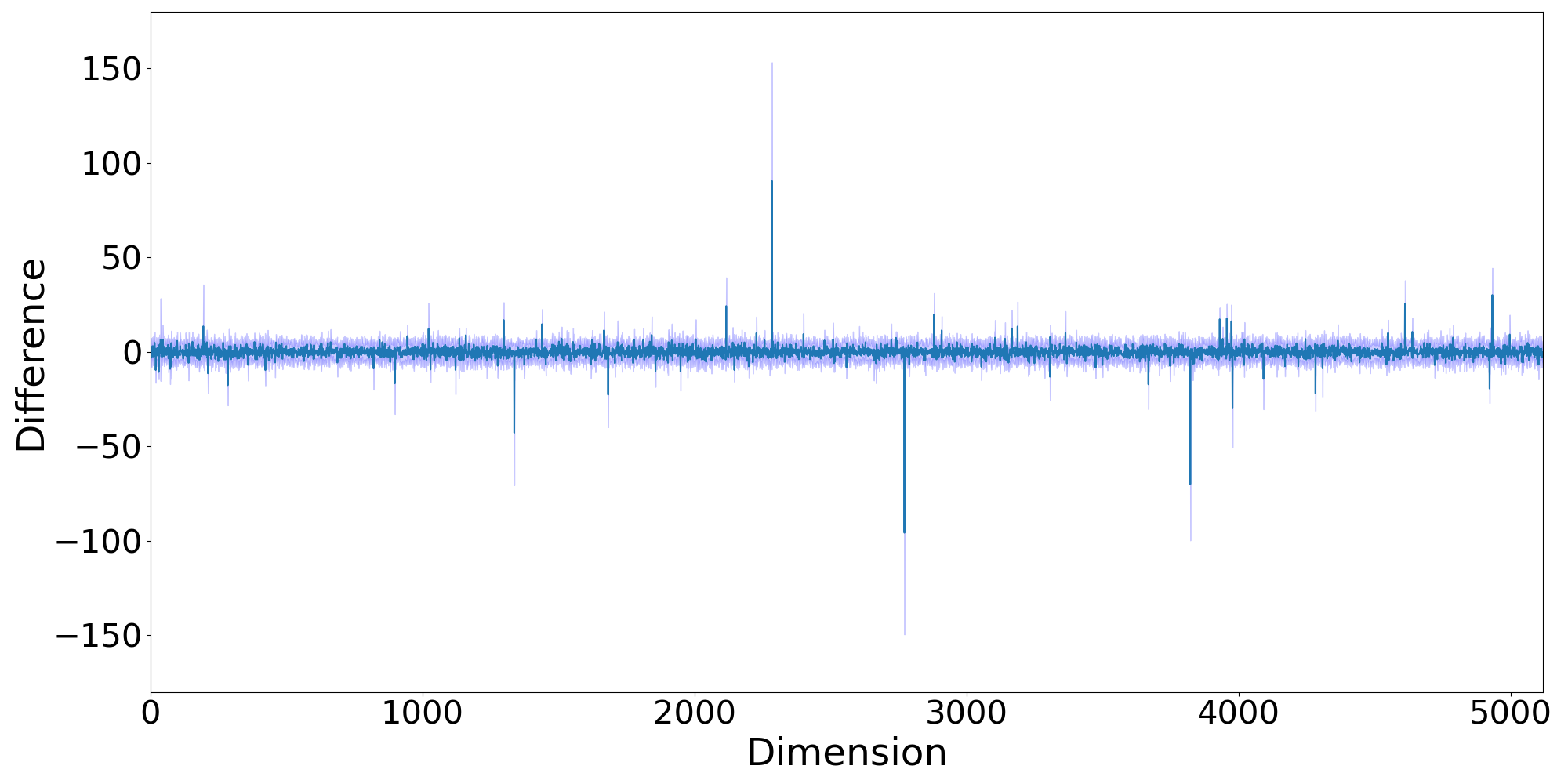}
    \caption{\textbf{Visualized dimensions of the average shift from layer 26 to 40 of mentioned in Figure~\ref{fig: how to shift}}. X-axis stands for the dimension IDs of the token presentation of 13B models and Y-axis shows the corresponding difference values between two layers.}
    \label{fig: shift}
\end{figure}

The process is shown in Figure~\ref{fig: how to shift}. In the 26th layer, the highest probability tokens are usually the English version of the word. After adding the shift, the top tokens become Japanese. In this way, one can directly approximate the output of the 40th layer. 
We then draw the average shift in a figure. As shown in Figure~\ref{fig: shift}, those substantial changes sparsely occurred in certain dimensions, which can be inferred to be related to language identity. Both results highly suggest that language identity dimensions can be distinguished serving the role of representing languages, while semantic dimensions are dense and shared across languages.

\section{Conclusion and Future Works}
In this study, we demonstrate that the internal latent language of LLMs is majorly determined by the language of its training corpora. We confirm that Japanese CPT model Swallow and trained bilingual from scratch model LLM-jp both utilize Japanese as their internal latent language when processing Japanese input. When dealing with languages that are not dominant in the pre-training corpora such as French and Chinese, both Swallow and LLM-jp exhibit the use of two internal latent languages. In both models, the internal latent language that is more closely related to the target language shows higher probability. For Swallow, the internal latent language distribution consistently includes both English and Japanese, with English having a higher probability. In contrast, LLM-jp tends to exhibit a more extreme favour towards one language.

Additionally, we observe that the transition from the internal latent language to the target language causes biased intermediate reasoning steps in the model's processes. We also discovered that the distribution transition within the latent layers occurs only in dense dimensions closely related to language identities. These findings provide valuable support for understanding why these models perform well in their specific dominant language and offer insights for future improvements.

In future research, we aim to extend our investigation to models with other specific dominant languages, such as Chinese, French, and Arabic, to further explore the behavior and mechanisms of non-English-centric LLMs.

\begin{CJK}{UTF8}{min}
\bibliography{custom}
\end{CJK}
\appendix

\section{Appendix}
\label{sec:appendix}

\begin{figure*}[t]
    \centering
    \begin{minipage}{\textwidth}
        \centering
        (a) Repetition: Zh
        \par\medskip
        \begin{minipage}{0.3\textwidth}
            \centering
            \includegraphics[width=\textwidth]{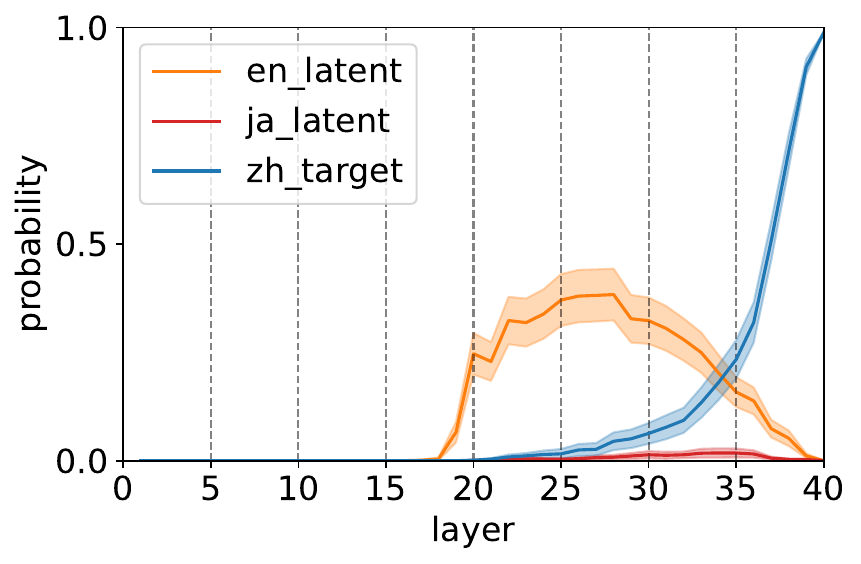}
        \end{minipage}
        \hfill
        \begin{minipage}{0.3\textwidth}
            \centering
            \includegraphics[width=\textwidth]{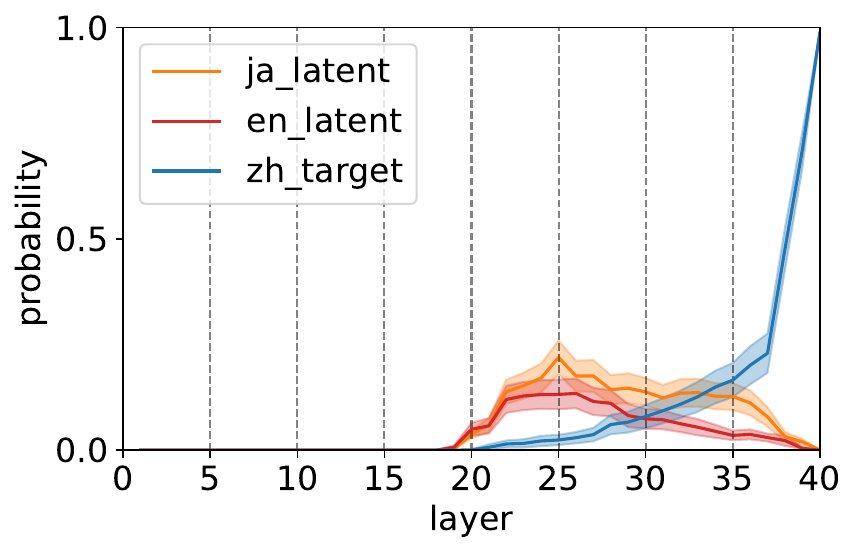}
        \end{minipage}
        \hfill
        \begin{minipage}{0.3\textwidth}
            \centering
            \includegraphics[width=\textwidth]{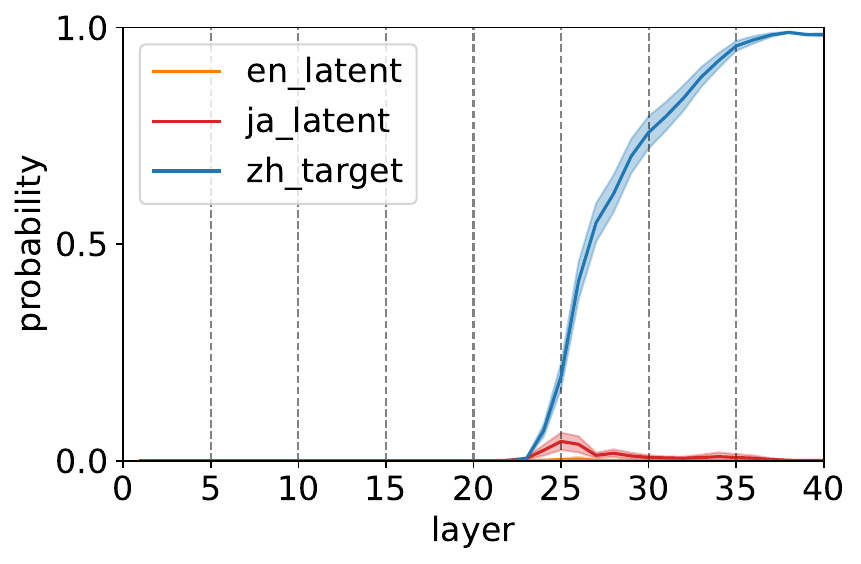}
        \end{minipage}
    \end{minipage}

    \par\bigskip

    \begin{minipage}{\textwidth}
        \centering
        (b) Repetition: Fr
        \par\medskip
        \begin{minipage}{0.3\textwidth}
            \centering
            \includegraphics[width=\textwidth]{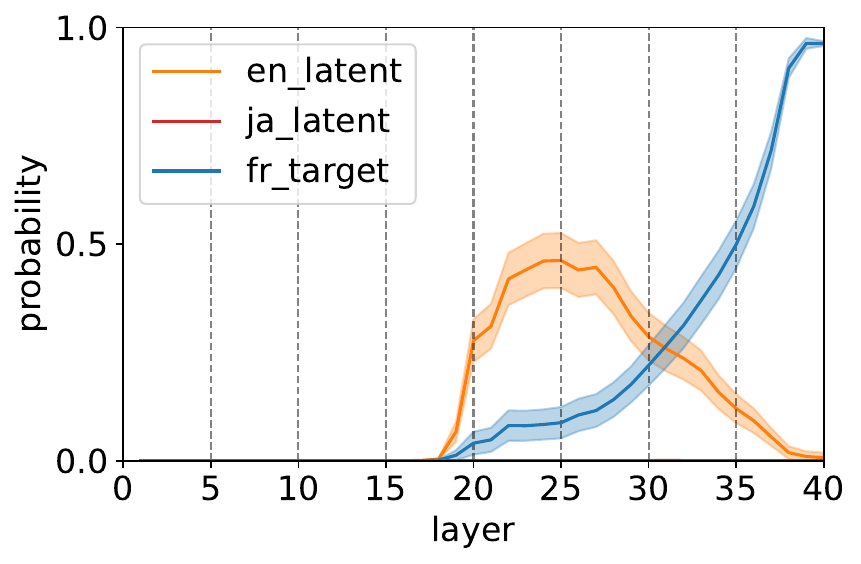}
            \par Llama-2-13b
        \end{minipage}
        \hfill
        \begin{minipage}{0.3\textwidth}
            \centering
            \includegraphics[width=\textwidth]{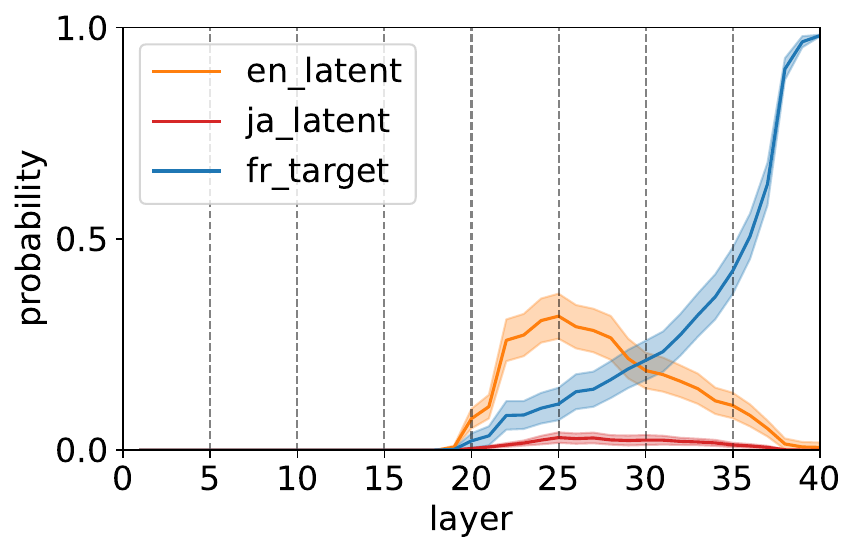}
            \par Swallow-13b
        \end{minipage}
        \hfill
        \begin{minipage}{0.3\textwidth}
            \centering
            \includegraphics[width=\textwidth]{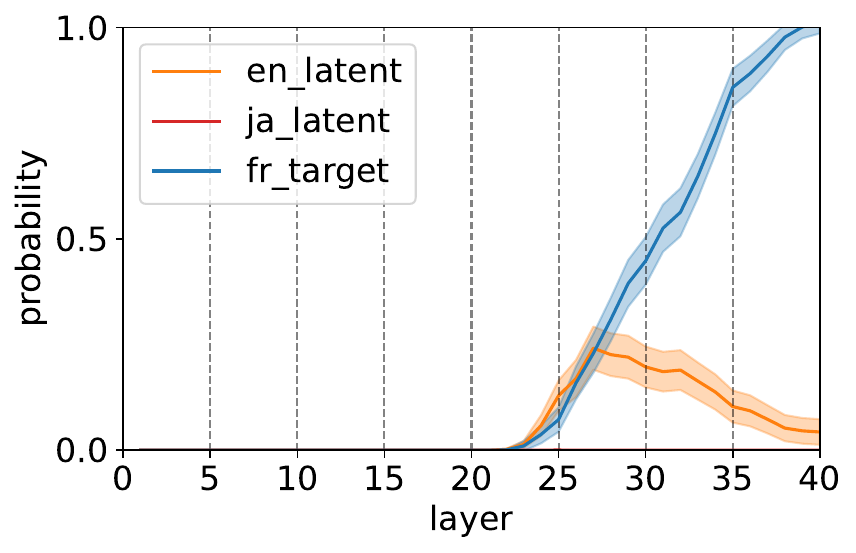}
            \par LLM-jp-v2.0
        \end{minipage}
    \end{minipage}
    \caption{\textbf{Language probabilities for three types of models doing repetition,} (a) repeat French, (b) repeat Chinese. X-axes stand for layer index and y-axes stand for probability of answer in each language. Error bars show 95\% Gaussian confidence intervals over totally 166 input examples.}
    \label{fig: repitition}
\end{figure*}

\begin{figure*}[t]
    \centering
    \begin{minipage}{\textwidth}
        \centering
        (a) Translation: Fr -> Zh
        \par\medskip
        \begin{minipage}{0.3\textwidth}
            \centering
            \includegraphics[width=\textwidth]{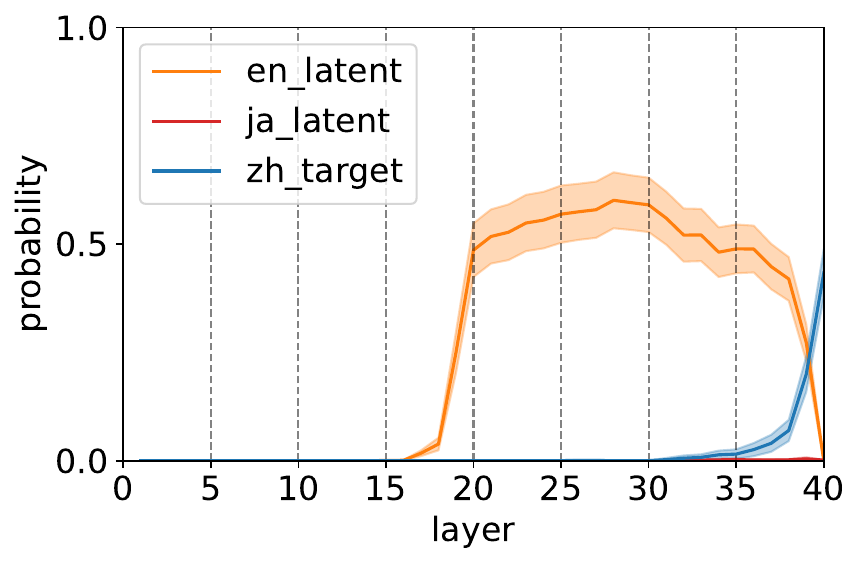}
        \end{minipage}
        \hfill
        \begin{minipage}{0.3\textwidth}
            \centering
            \includegraphics[width=\textwidth]{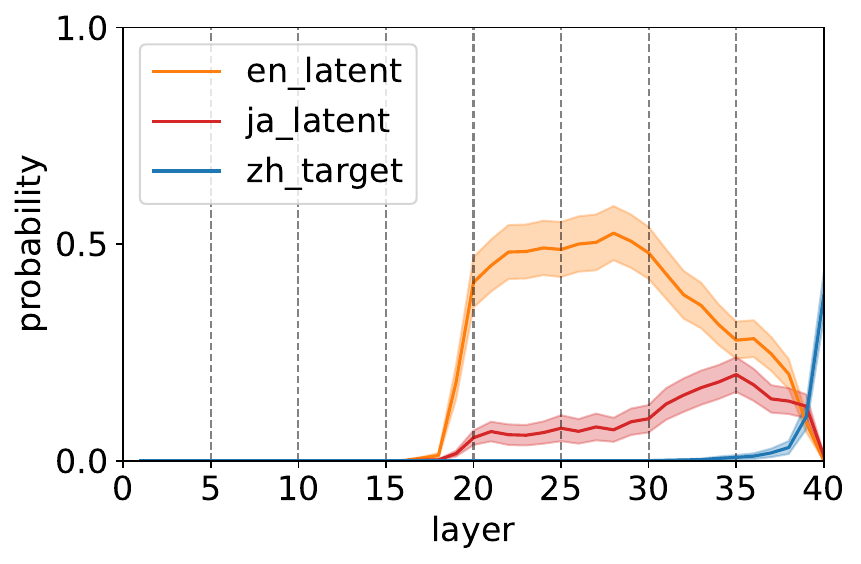}
        \end{minipage}
        \hfill
        \begin{minipage}{0.3\textwidth}
            \centering
            \includegraphics[width=\textwidth]{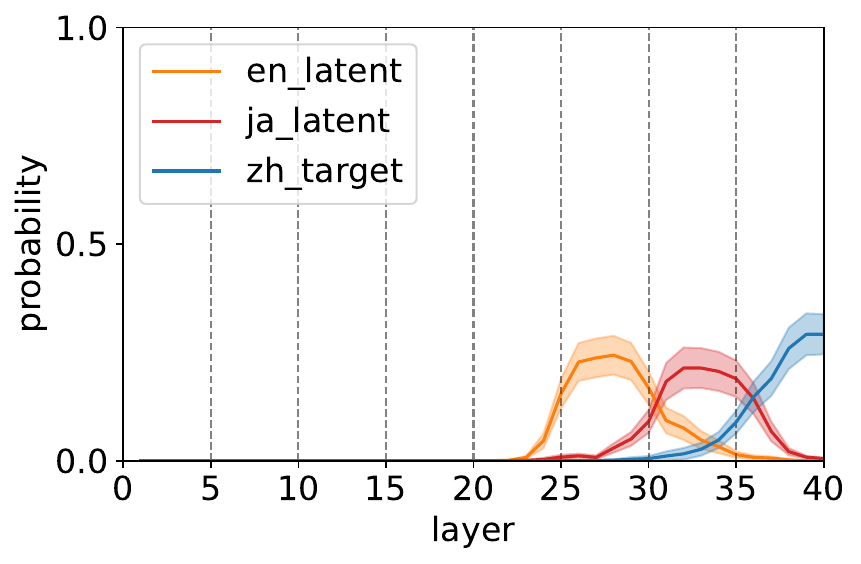}
        \end{minipage}
    \end{minipage}

    \par\bigskip

    \begin{minipage}{\textwidth}
        \centering
        (b) Translation: Zh -> Fr
        \par\medskip
        \begin{minipage}{0.3\textwidth}
            \centering
            \includegraphics[width=\textwidth]{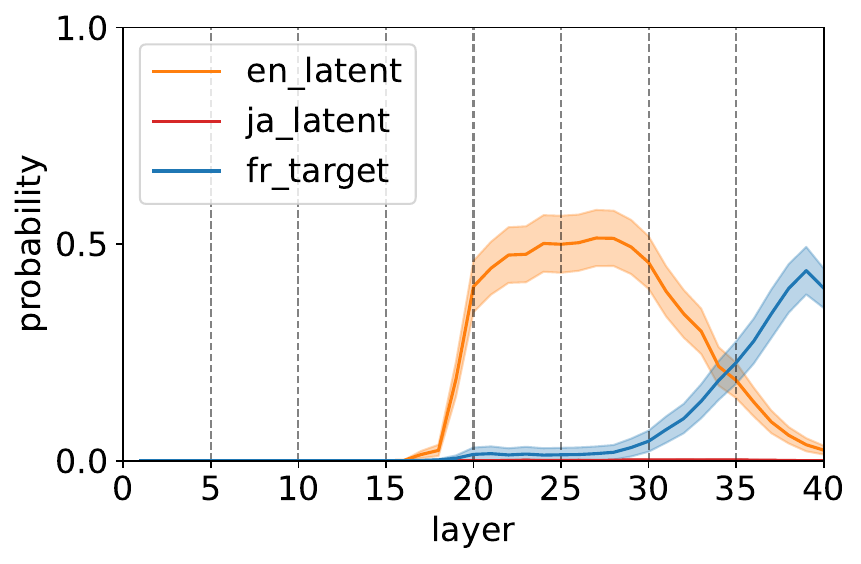}
            \par Llama-2-13b
        \end{minipage}
        \hfill
        \begin{minipage}{0.3\textwidth}
            \centering
            \includegraphics[width=\textwidth]{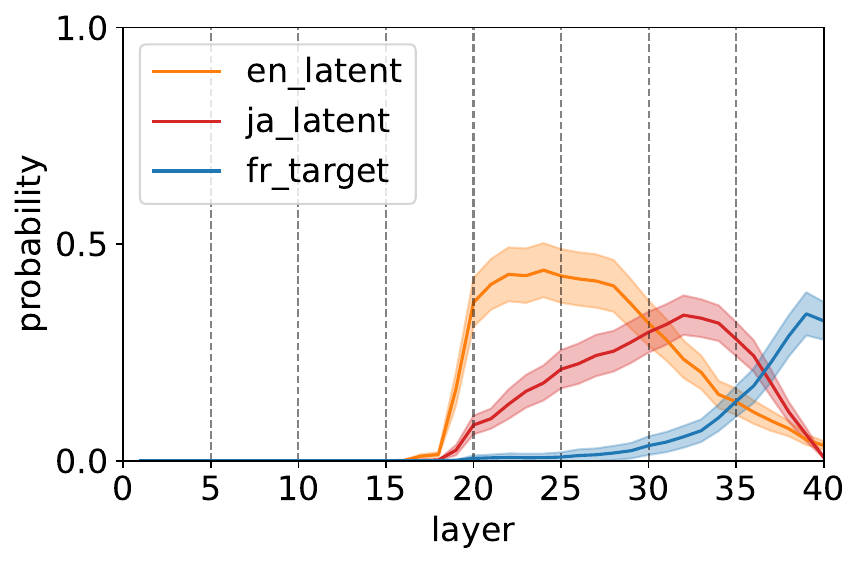}
            \par Swallow-13b
        \end{minipage}
        \hfill
        \begin{minipage}{0.3\textwidth}
            \centering
            \includegraphics[width=\textwidth]{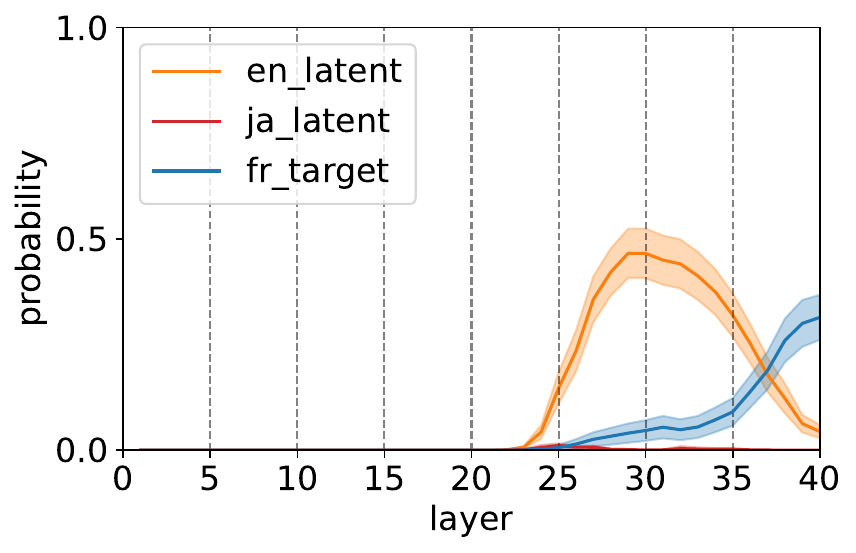}
            \par LLM-jp-v2.0
        \end{minipage}
    \end{minipage}
    \caption{\textbf{Language probabilities for three types of models in the translation task,} (a) French to Chinese, (b) Chinese to French. X-axes stand for layer index and y-axes stand for probability of answer in each language. Error bars show 95\% Gaussian confidence intervals over totally 166 input examples.}
    \label{fig: translation}
\end{figure*}


\end{document}